\documentclass{article}
\usepackage{spconf,amsmath,graphicx}

\usepackage{placeins}
\usepackage{amssymb}
\usepackage{rotating}
\usepackage{booktabs}
\usepackage{multirow}
\usepackage{verbatim}
\usepackage{lscape}
\usepackage{tabularx}
\usepackage{amssymb}
\usepackage{hyperref}
\usepackage{gensymb}
\usepackage{subfig}
\usepackage[svgnames,x11names,table]{xcolor}
\usepackage[belowskip=-6pt,aboveskip=0pt]{caption}
\usepackage{booktabs}

\usepackage{pdfpages}

\usepackage{adjustbox}
\usepackage{amsmath} 

\usepackage{nicefrac, xfrac}

\usepackage{url}
\newcommand{\mypar}[1]{\vspace{0.15em}
\textbf{#1}~}

\newcommand{\dorowcolors}{\rowcolors{2}{gray!15}{white}}

\usepackage{colortbl}
\usepackage{bbm}
\usepackage{bm,upgreek}
\usepackage{verbatim}

\newcommand{\mr}[1]{\mathit{#1}}

\newcommand{\fImg}{\mathbf{I}}
\newcommand{\fTxt}{\mathbf{T}}

\newcommand{\xx}{\mathbf{x}}

\newcommand{\real}{\mathbb{R}}

\newcommand{\PP}{\mathbf{P}}
\newcommand{\QQ}{\mathbf{Q}}
\newcommand{\SSS}{\mathbf{S}}

\usepackage{enumitem}
\setlist{nosep, leftmargin=14pt}

\usepackage{mwe} 

\title{Deep Modeling and Optimization of Medical Image Classification}
\name{Yihang Wu$^1$, Muhammad Owais$^1$, Reem Kateb$^{2}$, Ahmad Chaddad$^{1,3,*}$ }

\address{\parbox{0.99 \linewidth}{\centering $^1$AIPM, School of Artificial Intelligence, Guilin University of Electronic Technology, China\\
$^2$College of Computer Science and Engineering, Jeddah University, Jeddah, Saudi Arabia\\
$^3$Laboratory for Imagery, Vision and Artificial Intelligence, École de Technologie Supérieure, Canada\\ Correspondence:ahmad8chaddad@gmail.com
}}
\begin{document}
\maketitle

\begin{abstract}
Deep models, such as convolutional neural networks (CNNs) and vision transformer (ViT), demonstrate remarkable performance in image classification. However, those deep models require large data to fine-tune, which is impractical in the medical domain due to the data privacy issue. Furthermore, despite the feasible performance of contrastive language image pre-training (CLIP) in the natural domain, the potential of CLIP has not been fully investigated in the medical field. To face these challenges, we considered three scenarios: 1) we introduce a novel CLIP variant using four CNNs and eight ViTs as image encoders for the classification of brain cancer and skin cancer, 2) we combine 12 deep models with two federated learning techniques to protect data privacy, and 3) we involve traditional machine learning (ML) methods to improve the generalization ability of those deep models in unseen domain data. The experimental results indicate that maxvit shows the highest averaged (AVG) test metrics (AVG = 87.03\%) in HAM10000 dataset with multimodal learning, while convnext\_l demonstrates remarkable test with an F1-score of 83.98\% compared to swin\_b with 81.33\% in FL model. Furthermore, the use of support vector machine (SVM) can improve the overall test metrics with AVG of $\sim 2\%$ for swin transformer series in ISIC2018. Our codes are available at \url{https://github.com/AIPMLab/SkinCancerSimulation}. 

\end{abstract}
\vspace{-6pt}

\section{Introduction}
In recent years, deep learning models (DLMs) have significantly advanced medical imaging by using powerful architectures like convnext series, vision transformer (ViT) and maxvit for efficient deployment in realistic scenarios \cite{pacal2024maxcervixt}. These models have shown a remarkable ability to learn complex visual patterns and have outperformed traditional machine learning techniques in both accuracy and scalability. 

However, advanced DLMs require large amounts of source data to fine-tune, which is impractical in the field of healthcare due to data privacy concerns \cite{chaddad2023federated,wu2024facmic}. It can be challenging to collect and centralize the data required to effectively train algorithms since medical data is sensitive. Federated learning (FL), as a distributed learning framework that collaboratively trains a robust global model with multiple clients without sharing raw data, focuses on solving this challenge \cite{chaddad2023federated}. For example, federated averaging (FedAVG) can simply aggregate each client model with equal weights to produce the global model, demonstrating remarkable performance in many tasks \cite{mcmahan2017communication}. Furthermore, with the development of CLIP, multi-modal (e.g., image with text) demonstrates competitive performance in natural image classification \cite{radford2021learning}. However, a comprehensive evaluation using recent CNN or ViT models in the classification of skin cancer is still lacking. In addition, the authors in \cite{huix2024natural} also claimed that natural foundation models such as CLIP show poor testing recall in the ISIC2019 data set.

Motivated by previous challenges, we introduce two FL approaches to solve the data privacy leakage issue. Furthermore, we propose a CLIP variant (i.e., replace the image encoder with CNN or ViT) to evaluate the generalization ability of those deep models in medical imaging. This approach allows the model to learn comprehensive representations from both images and text, improving its ability to accurately classify skin cancer. Specifically, we perform extensive experiments using three common benchmark data sets including two modalities with 12 deep models (four CNNs, eight ViTs) to provide a comprehensive analysis. We summarize the key contributions of this approach as follows:

\begin{enumerate}
    \item We propose a novel CLIP-based approach that considers CNN and ViT architectures with CLIP text encoders for multimodal training, effectively integrating image and text data. 
    \item We introduce two FL approaches (FedAVG and FedProx) into skin cancer classification to solve data privacy leakage.
    \item To improve the generalization ability of deep models, we combine traditional ML with deep models. Specifically, we use ML techniques as classifiers while using deep models as feature extractors. 
\end{enumerate}

\begin{figure*}[]
    \centering
    \includegraphics[width=0.9\linewidth]{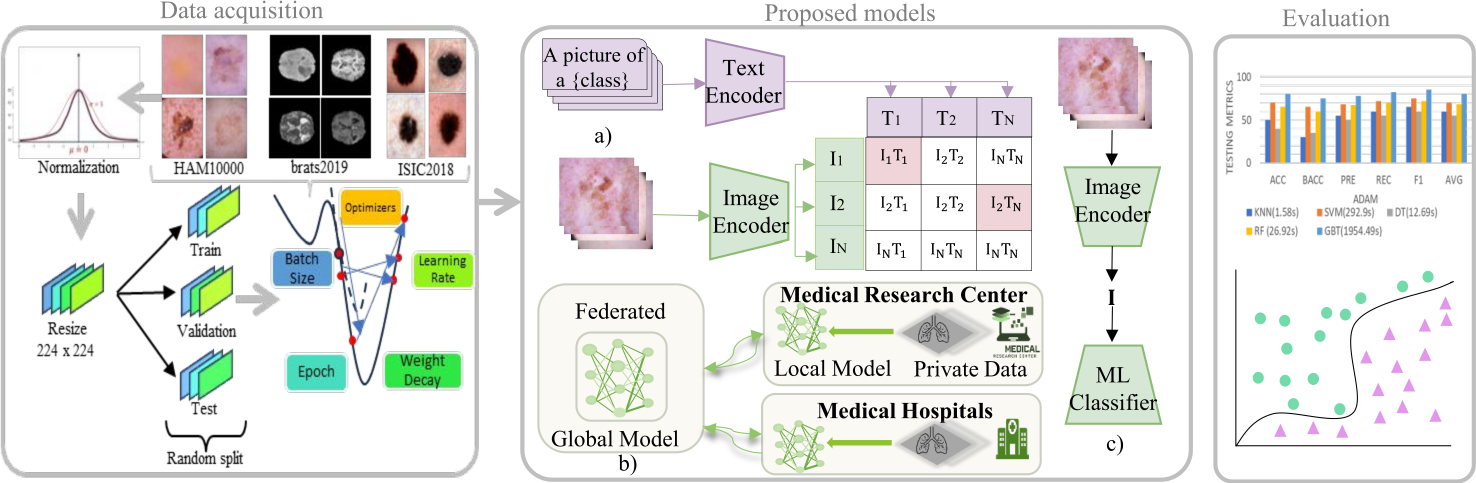}
    \caption{Flowchart of the proposed framework. 1) Data acquisition: Image data are preprocessed. 2) Proposed models: This involves three key components: multimodal learning, federated learning, and the combination of traditional ML and deep learning models. 3) Evaluation:  we evaluate those models performance using standard classification metrics.}
    \label{fig:flowchart}
\end{figure*}

\begin{figure*}[!ht]
    \centering
    \includegraphics[width=0.99\linewidth]{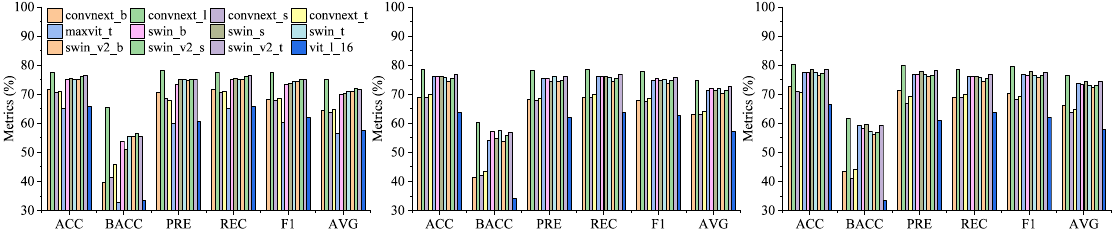}
    \caption{Test metrics on ISIC2018 dataset using deep models (\textbf{Left}), deep models with KNN (\textbf{Middle}) and deep models with SVM (\textbf{Right}).}
    \label{fig:LineChart_ISIC}
\end{figure*}

\begin{figure*}[!ht]
    \centering
    \includegraphics[width = 0.99 \linewidth]{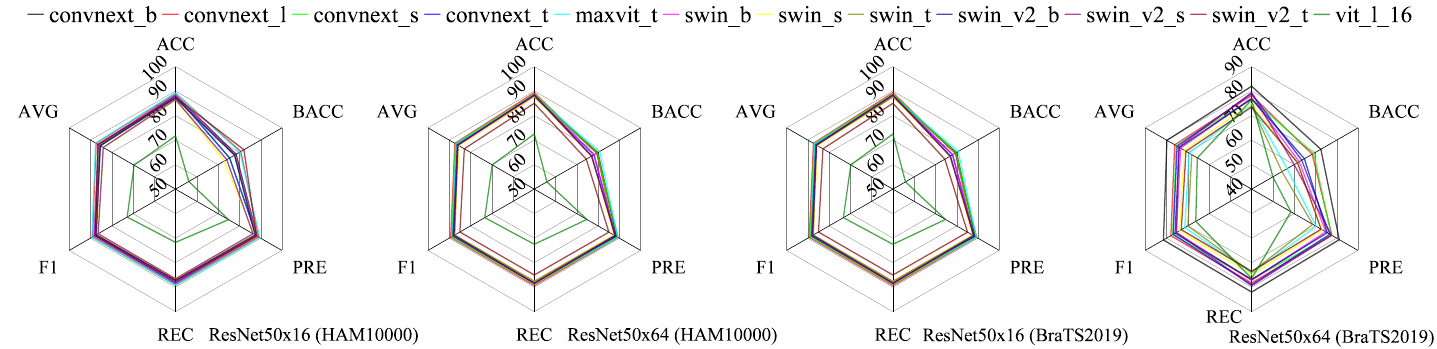}\\
    \includegraphics[width = 0.5 \linewidth]{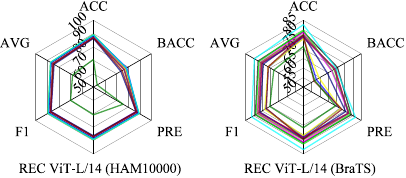}
    
    \caption{Spider-plot of multimodal test metrics (\%) for 12 deep models in HAM10000 and BraTS2019 using text encoder pretrained by three CLIP image encoders (ViT-L/14, ResNet50x16 and ResNet50x64).}
    \label{fig:Radar_CLIP}
\end{figure*}

\begin{figure*}[!ht]
    \centering
    \includegraphics[width = 0.99 \linewidth]{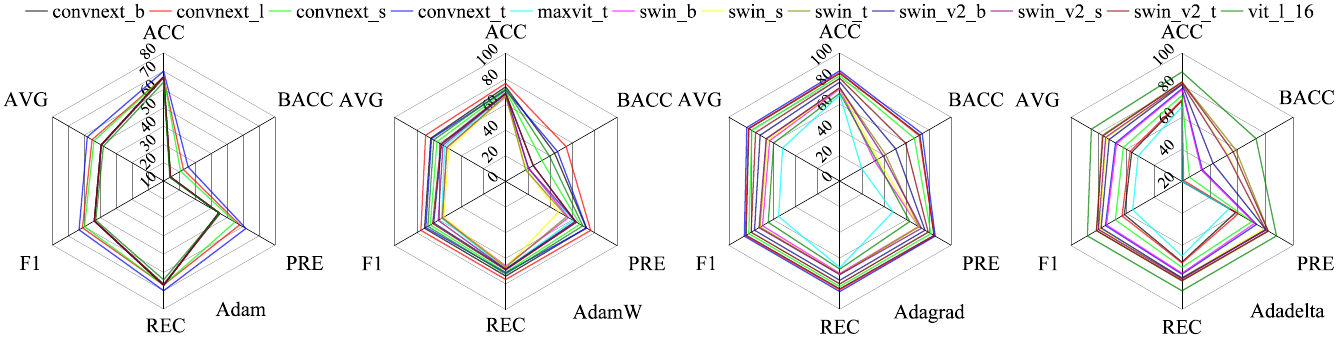}\\
\includegraphics[width = 0.99 \linewidth]{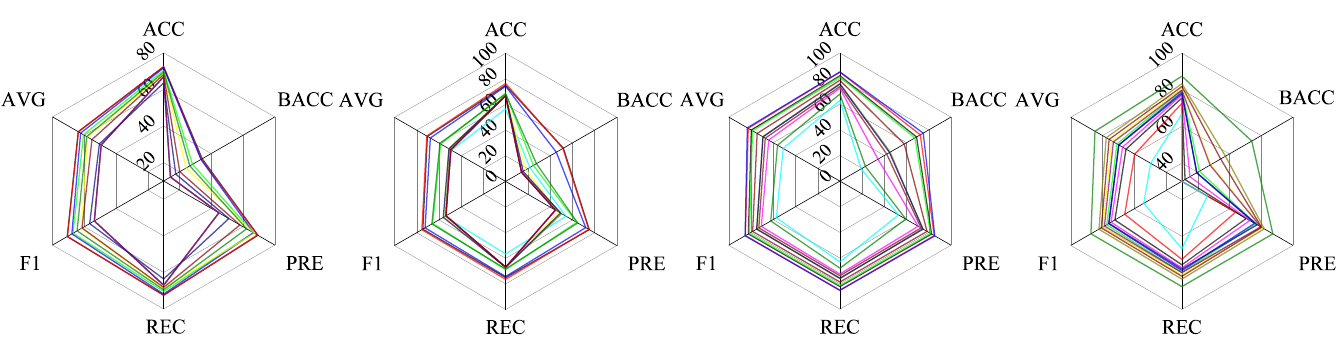}\\
\includegraphics[width = 0.5 \linewidth]{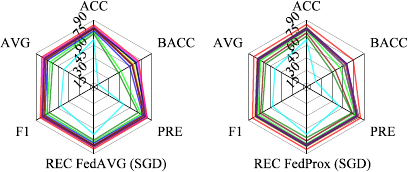}
    \caption{Spider-plot of global test metrics (\%) for 12 deep models in HAM10000 using FedAVG (First and third row) and FedProx (Second and third Row).}
    \label{fig:Radar_HAM_FL}
\end{figure*}

\vspace{-3pt}
\section{Methodology}
\vspace{-3pt}

Figure \ref{fig:flowchart} illustrates the flowchart of the proposed model. Specifically, we propose a CLIP variant based on CNN and VIT to extract the image feature and then combine it with the text encoder in CLIP to perform training and inference. In addition, we introduce two FL techniques with four CNNs and eight ViTs as network backbones to solve data privacy issue. Furthermore, we combine ML approaches with deep models to enhance the generalizability of deep classifiers. 

\noindent \mypar{Multimodal learning}
The key idea of CLIP is to maximize the similarities between paired image and text (e.g., class label) features while minimize the similarities between unpaired image and text features. However, a comprehensive analysis of using CLIP based deep models in skin cancer classification is still lacking. Motivated by this challenge, we propose to use these deep models to extract image features while using transformer to obtain text features to perform CLIP. Let $e_I(\cdot)$ be the image encoder while $e_T(\cdot)$ be the text encoder. For a training example $\xx_j$, we denote $\fImg_j = e_I(\xx_j) \in \real^D$ as the $D$-dimensional vector of image features. For text features, we use the standard prompt ``\texttt{a picture of a \{class\}}'' as input to the text encoder to obtain features $\fTxt_j = e_T(\xx_j) \in \real^D$. We calculate cosine similarity between the image and text features to measure the probability that $\xx_j$ belongs to a specific class $c$ as follows. 
\begin{equation}\label{eq:CLIP-classif}\footnotesize
p(\mathrm{Y}\!=\!c \, | \, \xx_j) \, =\,  \frac{\exp(s_{j,c}/\tau)}{\sum_{c'=1}^K \exp(s_{j,c'}/\tau)}, \ \ \text{with }
s_{j,c} = \frac{\langle\fImg^{*}_j, \fTxt_c\rangle}{\|\fImg^{*}_j\| \!\cdot\!\|\fTxt_c\|}
\end{equation}

Following \cite{radford2021learning}, the contrastive loss is used in to train the image encoder. Furthermore, we freeze the text encoder, only optimizing the image encoder. Let $B$ be the batch size, we compute the contrastive loss over batches of size $B$. Let $\SSS$ be the $B\!\times\!B$ matrix where $s_{j,j'}$ is the cosine similarity between the image features $\fImg^{*}_j$ and $\fTxt_{j'}$ as measured in Eq (\ref{eq:CLIP-classif}). We compute an image probability matrix $\PP = \mr{softmax}(\SSS/\tau) \in [0,1]^{B \times B}$ and a text probability matrix $\QQ = \mr{softmax}({\SSS}^{\textbf{T}}/\tau) \in [0,1]^{B \times B}$. The contrastive loss is then formulated as follows:
\begin{equation}\footnotesize
\mathcal{L}_{\mr{contr}} \, = \, -\frac{1}{B}\sum_{j=1}^B \frac{1}{2}\Big(\log p_{j,j} \, + \, \log q_{j,j}\Big). 
\end{equation}
where $p_{j,j}$ ($q_{j,j}$) is the vector in matrix $\PP$ ($\QQ$).

\noindent \mypar{Federated learning} In this study, we use two common FL techniques, FedAVG \cite{mcmahan2017communication} and FedProx \cite{li2020federated}. FedAVG is a FL methodology that using averaging aggregation to obtain the global model. FedProx adds a small proximal term to measure the discrepancy between the local and global model to enhance the generalizability of local models. After each local epoch, the clients will send the local models parameters (i.e., weights)  $\omega^{t}_{i}$ to the global server, while the global server will aggregate those weights by the aggregation techniques defined as follows.
\begin{equation}\footnotesize
    \omega^{t}_{glo} \,=\, \frac{1}{M}\sum\limits_{i = 1}^M {\omega _i^t}
\end{equation}
where $M$ is the number of clients.

Let $f_{\theta}$ be the global model, our goal is to train a robust global model that can perform well on each client test data, i.e.,
\begin{equation}\footnotesize
     \mathop {\min}\limits_f \ \frac{1}{N} \sum_{i=1}^N \frac{1}{n_i^{\mr{test}}} \sum_{j=1}^{n_i^{\mr{test}}} \mathcal{L}(f_\uptheta(\xx^{\mr{test}}_{i,j}),\, y^{\mr{test}}_{i,j}),
\end{equation}
where $\mathcal{L}$ is a given loss function, $\mathbf{x}^{\text{test}}_{i,j}$ is the $j$-th test input from client $i$, and $y^{\text{test}}_{i,j}$ is the corresponding label.



\noindent \mypar{Deep models with traditional machine learning} Deep models have remarkable ability of feature extraction, thereby using a simple linear layer can produce feasible performance in classification tasks. However, this may omit the merits of advanced traditional ML techniques such as Random Forest (RF). Inspired by \cite{almomani2024image}, we use traditional ML techniques as the classifier while using deep models as the feature extractor. These classifier models take the features extracted by deep models without the last classifier layer as inputs and predict based on these image features.
\begin{table}[!ht] \scriptsize
    \centering
    \renewcommand{\arraystretch}{1}
    \caption{Summary of hyper-parameter settings in optimizer.}
    \dorowcolors
    \setlength{\tabcolsep}{19pt}
    \begin{tabular}{cccc}
    \toprule
         &LR&WD&Betas \\
    \midrule
    SGD&0.01&0.0005&- \\
    Adam&0.001&0.02&(0.9,0.98)\\ 
    AdamW&0.001&0.02&(0.9,0.98)\\ 
    Adagrad& 0.001&0.0005&- \\
    Adadelta&0.001&0.0005&- \\
    \bottomrule
    \end{tabular}
    \label{tab:optimizers}
\end{table}

\vspace{-5pt}
\section{Experiments}
\vspace{-3pt}

\subsection{Datasets}

\mypar{Federated HAM10000 (FHAM).} We modified the HAM10000 dataset to build the FL dataset \cite{tschandl2018ham10000}. We split the original training set of HAM10000 into three clients with randomly selected samples, while the original testing set is used for global evaluation. In each client, the data are randomly partitioned into three non-overlapping parts, namely a training set (60\%), a validation set (20\%) and a testing set (20\%).

\noindent\mypar{ISIC2018.} ISIC2018 is a medium scale skin cancer dataset ($\sim$ 10000 images) with seven classes \cite{tschandl2018ham10000,codella2019skin}. For this dataset, we only considered its test set (1512 images) for the analysis of generalizability.

\noindent\mypar{BraTS2019.} We use a public kaggle dataset \cite{brats2019} with pre-divided data (7:1:2 for train, validation and test). It has two classes, namely high grade glioblastoma (HGG) and low grade glioma (LGG). The training set has 231 patients (178 HGG and 53 LGG), the validation set contains 32 patients (25 HGG and 7 LGG), while the test set holds 68 patients (52 HGG and 16 LGG).  

\subsection{Tasks}
\begin{enumerate}
    \item Multimodal learning. HAM10000 and BraTS2019 datasets are used to perform multimodal learning using 12 deep models (four CNNs and eight ViTs) with CLIP text encoders pretrained by three image encoders (i.e., ViT-L/14, ResNet50x16 and ResNet50x64). The Adagrad optimizer is considered for optimization.
    \item Federated skin cancer classification. We use FHAM dataset as an example to evaluate the usefulness of 12 deep models using two FL techniques (FedAVG and FedProx) with five optimizers.
    \item Generalization analysis. The ISIC2018 test set is used to demonstrate the generalizability of four CNNs and eight ViTs using k-nearest neighbours (KNN) and support vector machine (SVM) \cite{bansal2022comparative}. Note that the deep models are pretrained on HAM10000 training set using SGD optimizer (no overlap with ISIC2018 test set).
\end{enumerate}

\subsection{Implementation details}
We used convnext\_b, convnext\_l, convnext\_s, convnext\_t, maxvit\_t, swin\_b, swin\_s, swin\_t, swin\_v2\_b, swin\_v2\_s, swin\_v2\_t and vit\_l\_16 as the deep network backbones for classification tasks \cite{woo2023convnext,bangalore2024convision}. We choose SGD, Adam, AdamW, Adagrad, and Adadelta optimizers for simulations \cite{abdulkadirov2023survey}. Table \ref{tab:optimizers} reports a detailed hyper-parameter settings for those optimizers. The training epoch is set to 50. For \textbf{Task 1}, the batchsize is set to 16, while for \textbf{Task 2}, the batchsize is set to 32 for HAM10000 and 16 for BraTS2019. The batchsize of \textbf{Task 3} is set to 32. The experiment environment is based on the Windows 11 operating system and features an Intel 13900KF CPU with 128 GB of RAM and an RTX 4090 GPU. We use Pytorch 1.13.1 and Python 3.8. For classification metrics, this study considered accuracy (ACC), balanced accuracy (BACC), weighted-precision (PRE), weighted-recall (REC), weighted-F1 score and AVG (i.e., unweighted mean value of ACC, BACC, PRE, REC and F1).

\mypar{Task 1} Figure \ref{fig:Radar_CLIP} shows the test metrics in the HAM10000 and BraTS datasets. As illustrated, for Adagrad optimizer, maxvit\_t demonstrates the best overall testing metrics (87.03\% with ViT-L/14, 87.58\% with ResNet50x16) while convnext-large shows the highest overall testing metrics (87.36\% with ResNet50x64). This suggests that maxvit is more suitable to perform multimodal in skin cancer classification. Furthermore, unlike HAM10000, maxvit\_t fits well with the text encoder pretrained by ViT-L/14 (e.g., 79.6\% AVG) while performs poorly with the text encoder pretrained by ResNet50x64 (69.41\% AVG).

\mypar{Task 2} Figure \ref{fig:Radar_HAM_FL} shows the test metrics in HAM10000. As illustrated, convnext\_l demonstrates remarkable overall testing metrics (81.66\% using FedAVG and 82.72 using FedProx) with the SGD optimizer. Similarly, deep models with Adagrad optimizer indicate better performance compared to Adam, AdamW and Adadelta-based models. These findings suggest that the use of CLIP can achieve feasible performance, however, it introduces large communication costs in the FL context as suggested in \cite{hu2024federated}. In future work, we will explore efficient training techniques such as Adapter \cite{wu2024facmic}.

\mypar{Task 3} Figure \ref{fig:LineChart_ISIC} shows the metrics in the ISIC2018 test set. We observe that 1) introducing ML techniques such as KNN and SVM can improve the overall testing metrics of maxvit\_t $\sim 15\%$. 2) Similarly, the use of SVM indicates the best overall results compared to the original deep models. Those findings suggest that using ML techniques can further improve the performance of deep models in unseen domain.

\vspace{-3pt}
\section{Conclusion}
\vspace{-3pt}
This study proposed three models covering multimodal, FL and traditional ML with deep models in medical image classification tasks. The findings suggest that maxvit\_t shows potential for multimodal, convnext\_l indicates remarkable overall test metrics using FedAVG and FedProx with SGD optimizer, while introducing SVM and KNN can improve the overall performance of maxvit\_t, vit\_l\_16, convnext\_b, convnext\_l and swin transformer series in unseen domain. In future work, we will introduce domain adaptation \cite{chaddad2023enhancing} to minimize the data distribution shifts among different datasets to further improve deep models performance.

\section{Compliance with ethical standards}
\label{sec:ethics}
This is a numerical simulation study for which no ethical approval was required.

\vspace{-3pt}
\section{Acknowledgements}
\vspace{-3pt}
This research was funded by the National Natural Science Foundation of China \#82260360, the Guilin Innovation Platform and Talent Program \#20222C264164, and the Guangxi Science and Technology Base and Talent Project (\#2022AC18004 and \#2022AC21040).

\footnotesize
\bibliographystyle{ieeetr}
\bibliography{refs}

\end{document}